\documentclass{article}

\usepackage[preprint]{colm2026_conference}
\usepackage[T1]{fontenc}
\usepackage[utf8]{inputenc}
\usepackage{microtype}
\usepackage{inconsolata}
\usepackage{graphicx}
\usepackage{amsmath,amssymb}
\usepackage{booktabs}
\usepackage{multirow}
\usepackage{array}
\usepackage{tabularx}
\usepackage{capt-of}
\usepackage{url}
\usepackage{hyperref}
\definecolor{darkblue}{rgb}{0,0,0.5}
\hypersetup{colorlinks=true,citecolor=darkblue,linkcolor=darkblue,urlcolor=darkblue}
\graphicspath{{figures/}}

\newcommand{\hymemorycodeurl}{https://github.com/Tencent-Hunyuan/Hy-Memory}
\newcommand{\hymemorymodelurl}{https://huggingface.co/tencent/Hy-Memory-30B-A3B}
\newcolumntype{Y}{>{\raggedright\arraybackslash}X}
\renewcommand{\arraystretch}{1.08}
\newcommand{\paperlink}[3]{%
  \href{#2}{\raisebox{-0.22em}{\includegraphics[height=1.25em]{#1}}}%
  \hspace{0.55em}\href{#2}{\nolinkurl{#3}}%
}
\renewenvironment{abstract}{%
  \vskip.075in
  {\centering\large\bf Abstract\par}%
  \vspace{0.5ex}\noindent\ignorespaces
}{%
  \par\vskip 1ex
}

\title{%
\makebox[\textwidth][c]{%
\parbox{\textwidth}{\centering
Hy-Memory-30B-A3B: Attributed Reinforcement Learning for\\
Long-Term Memory Writing%
}}}

\author{\makebox[\dimexpr\textwidth-2\tabcolsep\relax][c]{%
  \begin{tabular}{c}
  Tianxiang Fei \qquad Jiazheng Kang \qquad Mao Zheng \\[0.25em]
  Xingyan He \qquad Xiang Yu \\[0.75em]
  Tencent Hunyuan
  \end{tabular}%
}}

\begin{document}
\AddToShipoutPictureFG*{%
  \AtPageUpperLeft{%
    \raisebox{-0.48in}[0pt][0pt]{%
      \hspace{0.85in}\includegraphics[height=0.27in]{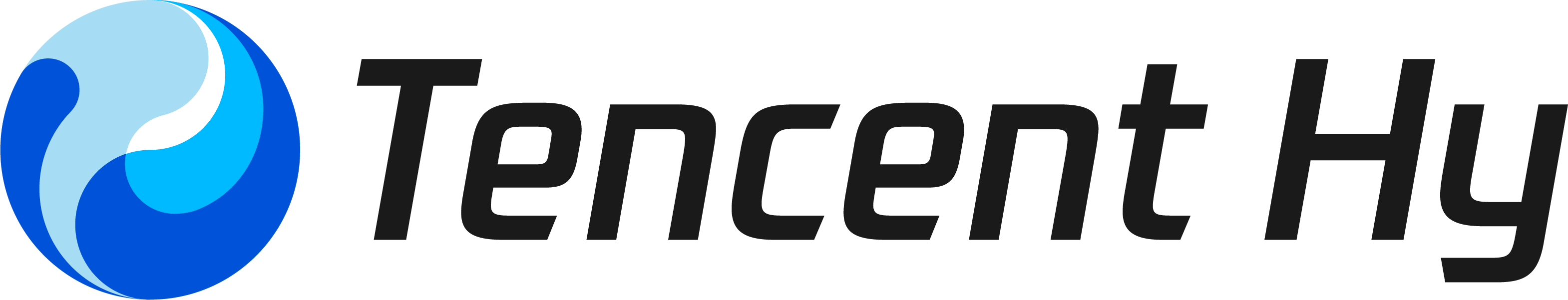}%
    }%
  }%
}
\maketitle
\lhead{}
\fancyheadoffset[L,R]{0.25in}

\begin{abstract}
External-memory systems extend language-model applications beyond a context window, but their quality depends critically on the model executing the write path: deciding what to retain, how precisely to represent it, and how new evidence should update persistent state. Existing frameworks typically place a general-purpose large language model behind framework-specific prompts. This raises serving cost, and a writer specialized for one protocol does not transfer directly to another; a unified model that can operate across established memory frameworks remains largely unexplored. We present \textbf{Hy-Memory-30B-A3B}, a compact long-term-memory writer trained across Mem0, MemOS, and Hy-Memory. Memory On-Policy Distillation teaches framework-specific write protocols on student-generated states; Memory RL then optimizes downstream answerability using dense per-chunk probes, attributed cross-chunk rewards, over-storage controls, and chunk-level GRPO. Carrying policy-written banks across windows keeps training aligned with deployment. On fixed 20\% benchmark subsamples, Hy-Memory-30B-A3B ranks first on all four evaluated harnesses, including held-out EverMemOS, with a 55.49\% macro average versus 54.62\% for Claude Opus~5.

\vspace{0.4em}
\begin{center}
\small
\begin{tabular}{@{}l@{}}
\paperlink{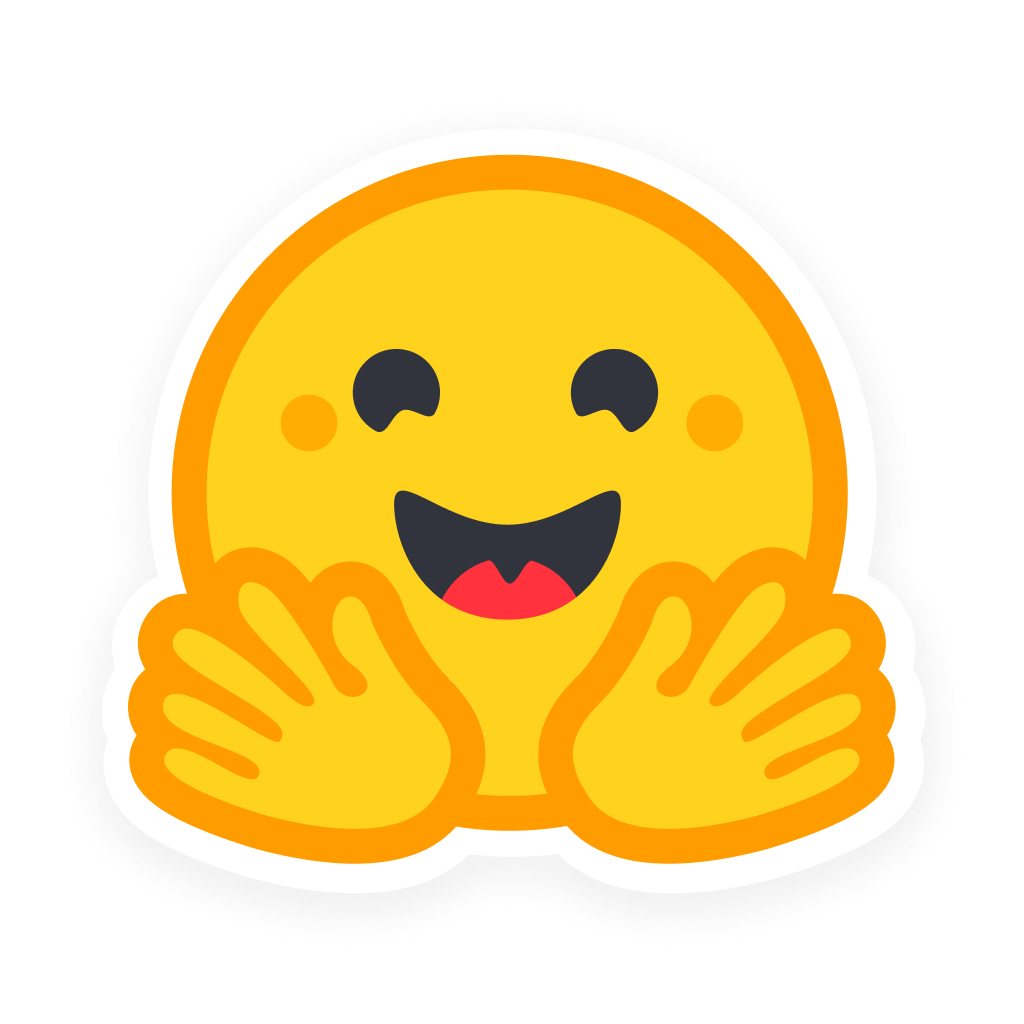}{\hymemorymodelurl}{huggingface.co/tencent/Hy-Memory-30B-A3B}\\[0.35em]
\paperlink{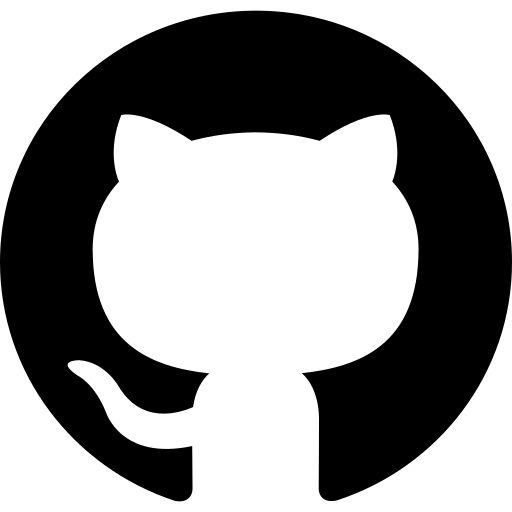}{\hymemorycodeurl}{github.com/Tencent-Hunyuan/Hy-Memory}
\end{tabular}
\end{center}
\end{abstract}

\section{Introduction}

A long-term memory system for an LLM application usually contains three functional stages:
(1)~a \textbf{writer} converts an interaction into persistent memory operations;
(2)~a \textbf{retriever} selects records relevant to a future query; and
(3)~a \textbf{reader} or answer model uses the retrieved records to respond.

Systems such as Mem0, A-MEM, MemoryOS, and MemOS have expanded the design space of persistent memory through extraction, consolidation, hierarchical paging, linking, versioning, and lifecycle management \citep{mem0,amem,memoryos,memos}. Benchmarks such as LoCoMo and LongMemEval have also shown that long-term memory is not a single retrieval problem: it includes precise extraction, multi-session composition, temporal reasoning, knowledge updates, and abstention \citep{locomo,longmemeval}.

Despite this progress, the write policy is commonly implemented by placing a general-purpose LLM behind a long, framework-specific prompt. That model must simultaneously distinguish durable user information from greetings, assistant suggestions, and transient text; preserve exact names, numbers, versions, dates, paths, and other identifying details; represent a state transition without losing either the new state or what it replaced; reconcile new evidence with retrieved memories already in the bank; and emit valid operations under the framework's schema and control flow.

The framework cannot recover information that the writer omitted, and a powerful downstream reader cannot reliably infer a fact that was never stored. Conversely, indiscriminately storing everything increases cost and retrieval interference. The desired writer is therefore not a generic summarizer: it is a policy for changing an external state under a long-horizon downstream objective.

Recent work has begun to train memory operations with reinforcement learning. Memory-R1 trains a memory manager and an answer agent with final-answer rewards \citep{memoryr1}. MemAgent learns recurrent textual compression from a terminal verifier \citep{memagent}. Mem-$\alpha$ combines downstream QA correctness with format, compression, and operation-quality rewards \citep{memalpha}. MemCoE explicitly identifies sparse outcome reward as a weakness and adds guideline-aligned process feedback \citep{memcoe}. These studies establish that memory behavior is learnable rather than fixed. They also expose a remaining problem: a final answer score usually supervises an entire sequence of heterogeneous memory actions, even though only a few actions supplied the evidence---or caused the failure.

Hy-Memory-30B-A3B focuses on this credit-assignment problem while preserving compatibility with existing memory frameworks. Our contributions are:
\begin{itemize}
\item \textbf{A framework-integrated memory writer.} Hy3-30B-A3B replaces the LLM inside Mem0, MemOS, and Hy-Memory without moving their memory logic into a custom model architecture.
\item \textbf{State-distribution-aligned training.} OPD and RL both execute the policy inside the actual harness, and later writes observe banks produced by earlier policy outputs.
\item \textbf{Failure-surface reward data.} Training questions are constructed from concrete write failures, including omission, lost specificity, semantic inversion, stale-state confusion, secondary-fact loss, and over-storage.
\item \textbf{Attributed, non-binary feedback.} Local probes provide dense per-chunk evidence, while cross-chunk outcome questions use gold facts to map recovered evidence back to source chunks.
\item \textbf{Chunk-level policy optimization.} Advantages are normalized among rollouts at the same chunk position rather than over whole trajectories, so one rollout can be credited at one chunk and penalized at another. Each rollout casts one vote per chunk even when a framework makes multiple LLM calls.
\item \textbf{Cross-harness and OOD results.} Hy-Memory-30B-A3B ranks first on Mem0, Hy-Memory, MemOS, and held-out EverMemOS, averaging 55.49\% across the four harnesses versus 54.62\% for Claude Opus~5.
\end{itemize}

\section{Background and Related Work}

\subsection{External memory systems and evaluation}

Mem0 uses incremental extraction and consolidation to maintain a persistent memory store \citep{mem0}. A-MEM organizes memories as dynamically linked notes whose attributes and relationships evolve as new information arrives \citep{amem}. MemoryOS introduces an operating-system-inspired hierarchical storage architecture that coordinates short-term, mid-term, and long-term personal memory through dialogue-chain FIFO and segmented page updates \citep{memoryos}. MemOS treats memory as a managed system resource and emphasizes representation, scheduling, provenance, and lifecycle control \citep{memos}. These systems differ in schema and control flow, but all depend on an LLM to interpret new interactions and produce valid operations.

DCPM argues that long-term memory should support more than surface recall: its dual-process architecture represents belief revisions, identity, schemas, latent intentions, and cross-domain abstractions in a capability hierarchy \citep{dcpm}. This direction is complementary to ours. DCPM redesigns the memory architecture and its evolution process, whereas Hy-Memory-30B-A3B trains a reusable writer that operates inside several existing memory architectures without replacing them.

LoCoMo evaluates conversations containing hundreds of turns and multiple sessions, with questions requiring single-hop, multi-hop, temporal, open-domain, and adversarial reasoning \citep{locomo}. LongMemEval decomposes conversational memory into information extraction, multi-session reasoning, temporal reasoning, knowledge update, and abstention \citep{longmemeval}. Together, these benchmarks motivate evaluating the full memory pipeline rather than the fluency of an extracted summary.

Our work draws two conclusions from this literature. First, memory quality must be measured downstream, after retrieval. Second, a useful reward dataset must cover distinct failure modes rather than collapsing all memory behavior into one generic QA score.

\subsection{Reinforcement learning for memory}

There are two related but distinct lines of memory RL.

\paragraph{Recurrent compression agents.}
MemAgent processes a long document in chunks, repeatedly overwrites a bounded textual memory, and applies a final verifiable answer reward to all generated conversations \citep{memagent}. This trains a model-native long-context mechanism, but the memory representation and inference loop are part of the learned architecture.

\paragraph{External-bank management.}
Memory-R1 learns ADD, UPDATE, DELETE, and NOOP operations, and also trains an answer agent that filters retrieved memories \citep{memoryr1}. Mem-$\alpha$ formulates memory construction as a sequence of operations over a multi-component bank and combines global QA correctness with local format and semantic-validity rewards \citep{memalpha}. MemCoE uses an induced guideline to construct denser process feedback because final answer rewards alone are sparse over long memory-evolution trajectories \citep{memcoe}.

Hy-Memory-30B-A3B belongs to the second line but differs in scope. It trains the \textbf{writer only}, keeps the reader fixed during reward computation, and targets multiple pre-existing harnesses rather than one newly designed memory architecture. If the writer and answer model are optimized together, a better answer can come from improved reading rather than improved memory. Table~\ref{tab:related} summarizes the distinction.

State separation also appears outside memory-specific agents. CodeDelegator separates persistent orchestration state from ephemeral execution sandboxes to prevent debugging traces and intermediate failures from polluting long-horizon planning context \citep{codedelegator}. Our setting has a related systems requirement: candidate rollouts must write to isolated banks, while only the selected bank persists across windows. Unlike CodeDelegator, however, our separation is used for on-policy credit assignment over memory writes rather than role-specialized code execution.

\begin{table}[t]
\centering
\small
\setlength{\tabcolsep}{4pt}
\begin{tabularx}{\linewidth}{@{}p{2.0cm}YY@{}}
\toprule
Method & Learned policy and environment & Reward and distinction \\
\midrule
Memory-R1 & Manager and answer agent over a structured external bank & Downstream answer outcome; writer is confounded with the learned reader \\
MemAgent & Recurrent writer and answer policy over model-native fixed memory & Terminal score broadcast over the trajectory; not a drop-in writer \\
Mem-$\alpha$ & Construction policy over a custom multi-component bank & Global QA plus local checks; global correctness is shared across actions \\
MemCoE & Guideline and evolution policy for personalization memory & Process plus final-answer reward; checks guideline adherence rather than evidence contribution \\
\textbf{Hy-Memory-30B-A3B} & Writer only, operating in Mem0, MemOS, and Hy-Memory & Per-chunk recall, attributed outcome, and growth penalty; localizes evidence to writes \\
\bottomrule
\end{tabularx}
\caption{How Hy-Memory-30B-A3B differs from prior memory RL.}
\label{tab:related}
\end{table}

\subsection{Why ordinary outcome-only RL is insufficient}

LLM RL commonly assigns a scalar reward to a whole completion. PPO stabilizes policy updates with a clipped surrogate \citep{ppo}, while GRPO replaces a learned critic with a baseline estimated from grouped samples \citep{grpo}. This works well when the completion is a single solution and the final verifier is informative. A memory trajectory violates both assumptions.

A write in an early session may matter only when a much later question asks about it, so most writes receive no direct evidence. Giving the final QA score to every write rewards irrelevant actions and penalizes correct writes that co-occur with one bad action. Binary exact match often makes every rollout in a small GRPO group receive the same score, producing zero advantage and no task gradient; this has also been observed in memory-grounded QA \citep{he2026}. Jointly improving the answer agent can raise QA accuracy while the bank remains incomplete or noisy. An action changes the bank seen by later actions, so examples are not independent prompts. One write may invoke one LLM call while another invokes several, so a row-level baseline can weight call count rather than decision quality. Recall-only rewards can favor excessive storage.

The distinction between outcome and process supervision is well established in reasoning \citep{lightman}. Memory writing needs a related but domain-specific decomposition. We do not label a hidden reasoning chain. Instead, we supervise observable consequences of each bank mutation: whether the introduced fact remains recoverable, whether irrelevant text was stored, and whether later cross-chunk answers recover the evidence contributed by that chunk.

\section{Problem Formulation}

\subsection{Unified harness scope}

OPD and RL use the same conceptual harness set (Table~\ref{tab:harness}). The three Hy-Memory rows are scenario configurations of one harness family, not three unrelated frameworks. ``All trained harnesses'' means these five configurations from three harness families.

\begin{table}[t]
\centering
\small
\setlength{\tabcolsep}{5pt}
\renewcommand{\arraystretch}{1.12}
\begin{tabularx}{\linewidth}{@{}p{1.95cm}p{2.05cm}Y@{}}
\toprule
Family & Configuration & Role \\
\midrule
Mem0 & General & Incremental extraction and additive writing \\
MemOS & General & Multi-stage extraction and reconciliation \\
Hy-Memory & Chat & Personal conversational memory \\
Hy-Memory & Working & Tasks, decisions, and project state \\
Hy-Memory & Coding & Code, commands, bugs, and technical decisions \\
\bottomrule
\end{tabularx}
\caption{Harness families and scenario configurations.}
\label{tab:harness}
\end{table}

\subsection{Data hierarchy}

An interaction sequence contains chronological sessions. Each session contains turns (one request and its response). A \emph{chunk} is one harness write input, usually about ten turns, and never crosses a session boundary. A \emph{window} is up to five consecutive chunks processed in one RL step and may span session boundaries.

\subsection{Memory writing as an MDP}

For chunk $t$, let $M_t$ be the persistent memory bank and $x_t$ the raw interaction chunk. The policy receives a harness-generated prompt containing $x_t$ and selected information from $M_t$, then produces one or more structured LLM outputs $a_t$. The harness executes those outputs and creates the next bank:
\begin{align}
s_t &= (M_t,x_t,h), \nonumber \\
a_t &\sim \pi_\theta(\cdot\mid s_t), \quad
M_{t+1}=T_h(M_t,a_t),
\end{align}
where $h$ denotes the harness configuration. The state lives primarily in the external bank, not in a continuous model context. Every LLM call may start from a fresh prompt, but the decisions are sequential because the bank carries their consequences.

This creates two training requirements: rollouts from one candidate must write to an isolated bank copy, and later windows must start from a bank the policy itself produced, not an annotation-derived ideal bank.

\subsection{Three-model separation}

The policy (Hy3-30B-A3B) sees the harness write prompt and produces all memory-write outputs; it is the only model trained in Stage~2. The answer model sees the question and the retrieved memory block, and reads what the bank exposes. The judge sees the question, answer, reference, and rubric, and assigns semantic credit. The answer model never sees the reference or rubric. The judge never sees the retrieved memory block. Freezing both makes the memory bank the only changing component in the reward path.

\section{Two-Stage Training}

\subsection{Why two stages}

Direct RL from a generic instruction model wastes samples on protocol failures: malformed JSON, wrong operation names, missing fields, and invalid multi-stage control flow. Conversely, pure distillation teaches the model to follow a strong teacher but cannot rank the many valid memory representations by downstream usefulness. OPD learns how to act inside each harness, with dense token-level supervision on the student's own prompts and outputs. RL learns which valid actions produce useful banks.

\begin{figure}[t]
\centering
\includegraphics[width=\linewidth]{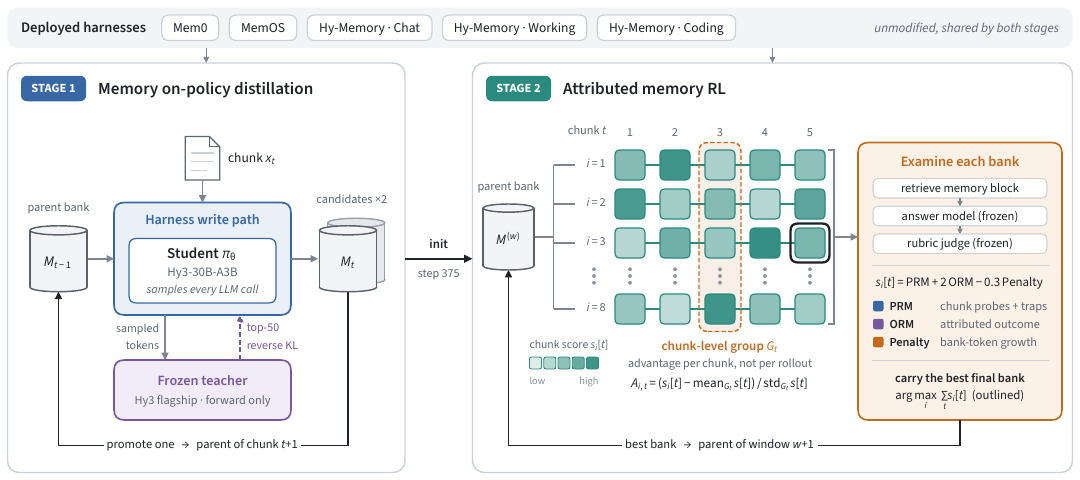}
\caption{Hy-Memory-30B-A3B is trained inside the same deployed harness configurations in both stages. OPD supplies dense teacher supervision on student-written state chains; RL examines the resulting banks, assigns chunk-level credit, and carries the selected bank across windows. In the rollout grid, each cell is the bank after chunk $t$ of rollout $i$, shaded by its chunk score $s_i[t]$. Advantages are normalized within each column (dashed: one chunk-level group), so the best rollout can differ from chunk to chunk. The outlined cell is the final bank with the highest $\sum_t s_i[t]$, which becomes the parent of the next window.}
\label{fig:train}
\end{figure}

\subsection{Stage 1: Memory on-policy distillation}

\paragraph{Sequence construction.}
Chronologically ordered interaction sequences are segmented into chunks while preserving session boundaries. Chat emphasizes personal facts and preference changes; Working emphasizes tasks, decisions, deadlines, and organizational context; Coding emphasizes repositories, files, commands, implementation decisions, and changing technical state. Each chunk is paired with the harness state produced by preceding student writes, so OPD supervision follows the state distribution encountered during deployment.

\paragraph{On-policy rollout.}
Each OPD row advances one (sequence, harness configuration) snapshot chain by one chunk. The rollout restores the selected parent bank $M_{t-1}$, invokes the unmodified harness write path, routes every harness LLM call to the student, applies the student's operations to produce a candidate $M_t$, and promotes one candidate as the parent of the next chunk.

The teacher does not generate a separate ideal trajectory. It scores tokens on the trajectory the student generated. Supervision is placed on student-visited prefixes, reducing the exposure mismatch of static teacher demonstrations \citep{minillm}. For memory writing, the benefit extends beyond token prefixes: the future prompts also contain memory states created by the student.

\paragraph{Objective.}
The frozen Hy3 flagship teacher runs a forward pass on the student's token stream and returns its top-$k$ distribution. With $k=50$, the response-token objective is a top-$k$ approximation to reverse KL:
\begin{equation}
\begin{aligned}
\mathcal{L}_{\mathrm{OPD}}
&= \frac{1}{|\mathcal{T}|}
   \sum_{(i,j)\in\mathcal{T}} \ell_{i,j}, \\
\ell_{i,j}
&= \sum_{v\in\mathrm{top}\text{-}50}
   \pi_\theta(v\mid x_{i,<j})
   \left[
   \log\pi_\theta(v\mid x_{i,<j})
   -\log\pi_T(v\mid x_{i,<j})
   \right].
\end{aligned}
\end{equation}
MiniLLM similarly motivates reverse KL on student-sampled trajectories \citep{minillm}. All LLM calls made by one harness write are retained as explicit segments. Position ids reset at call boundaries, so independent calls do not become one artificial conversation.

\paragraph{Configuration.}
The student is the Hy3-30B-A3B instruction-tuned checkpoint ($\sim$3B active parameters). The teacher is the Hy3 flagship model, forward only, returning top-50 log-probabilities. There are 2 rollouts per row, learning rate $1\mathrm{e}{-6}$ (constant), and entropy coefficient $0.005$. Prompt and response limits are 4{,}096 and 126{,}976 tokens across segmented calls. The checkpoint that initializes RL is OPD step 375. The unified configurations are Mem0, MemOS, and Hy-Memory Chat, Working, and Coding.

\subsection{Stage 2: Memory reinforcement learning}

RL starts from OPD step 375 and executes the same harness configurations. For each window, eight rollouts begin from independent copies of the same parent bank. They write all chunks, are examined with reward questions, and receive per-chunk scores. The highest-scoring final bank is promoted to the next window in the same sequence. This is on-policy at two levels: tokens are sampled from the current writer, and future prompts are generated from banks produced by that writer.

\section{Reward Construction: From Failure Surfaces to Attributed Feedback}

\subsection{Why benchmark questions are not enough}

Using only final questions from a long-memory benchmark creates a low-density learning problem. A long interaction sequence may contain hundreds of useful facts but only a few questions. An unasked write receives no evidence, and a failed answer does not identify whether the cause was omission, bad wording, stale-state confusion, retrieval failure, or reader failure. Binary exact match further collapses semantically partial answers and can eliminate variance within a small GRPO group \citep{he2026}. We construct reward items from the failure surface of memory writing, not merely from a list of benchmark answers.

\subsection{Chronological segmentation}

Interaction sequences retain session boundaries and chronological order. Each session is segmented independently into chunks of roughly ten turns, so a chunk never crosses a session boundary. The resulting chunks are flattened into one chronological stream; up to five consecutive chunks form one RL window, which may span sessions. Windows from the same sequence are consumed in order, and the selected policy-written bank from one window initializes the next. This preserves long-range state transitions while keeping each optimization step bounded.

\subsection{Two annotation passes}

Local write quality and long-range memory utility require different context, so the two reward types are annotated in separate passes (Table~\ref{tab:passes}).

\begin{table}[t]
\centering
\small
\setlength{\tabcolsep}{4pt}
\begin{tabularx}{\linewidth}{@{}p{2.2cm}YY@{}}
\toprule
Pass & Annotator context & Output and target \\
\midrule
Per-chunk probe (PRM) & Full text of chunk $t$ plus facts of chunks $t{-}1$ and $t{-}2$ & Atomic facts, probes, traps, and rubrics; tests whether chunk $t$ was written precisely and remains recoverable \\
Cross-chunk outcome (ORM) & Earlier-sequence facts plus all facts in the current window & Questions, evidence chunks, gold facts, and rubrics; tests relations spanning chunks or sessions \\
\bottomrule
\end{tabularx}
\caption{The two annotation passes and the reward each one feeds.}
\label{tab:passes}
\end{table}

\paragraph{Why the PRM pass sees the two preceding chunks.}
A per-chunk probe must test information that chunk $t$ itself introduced. Annotating chunk $t$ in isolation fails in two ways. \emph{Coreference:} mentions such as ``that project'' or ``the same reviewer'' cannot be resolved, so the annotator cannot write self-contained questions about them. \emph{Novelty:} the annotator cannot tell whether a statement is new in chunk $t$ or repeats something already established, so probes would credit chunk $t$ for facts it did not introduce. The extracted facts of chunks $t{-}1$ and $t{-}2$ supply the referents and the recently established state. They are given as facts rather than raw text and serve as context only: no probe is generated from them, and they earn no credit for chunk $t$.

\paragraph{Why not the full history.}
Dependencies that span more than a few chunks are covered by the cross-chunk outcome pass, so the PRM pass does not need the whole sequence. Full history also lowers annotation quality. In a controlled comparison on identical chunks, full-history annotation used 33 times more input tokens, while the two-chunk context produced 25\% more proper nouns and 59\% more numbers per extracted fact. With the full sequence in view, the annotator's attention spreads away from the current chunk and its facts become more generic, which is the over-generalization failure (Section~\ref{sec:taxonomy}) that probes are meant to detect.

\begin{figure}[t]
\centering
\includegraphics[width=\linewidth]{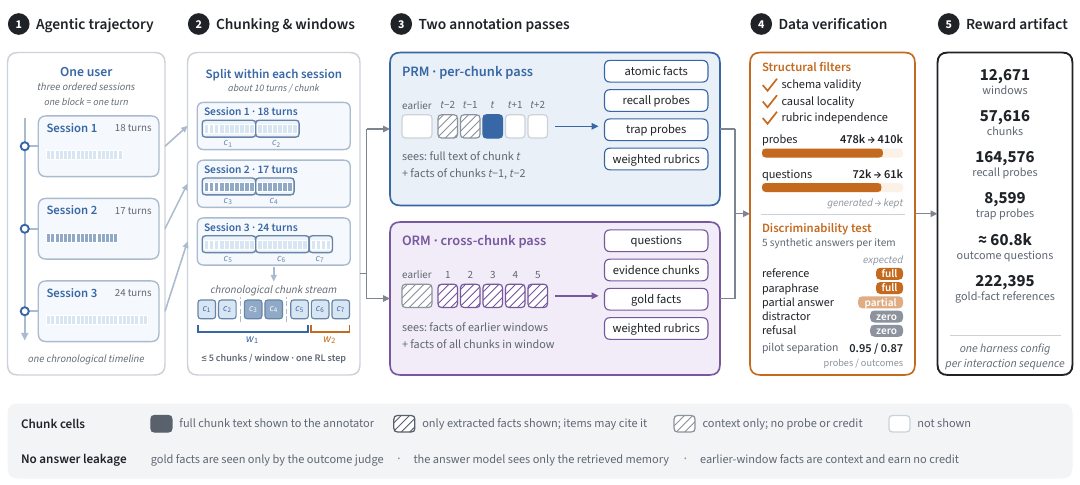}
\caption{Construction of the Memory RL reward artifact. Each session is split independently into chunks of about ten turns, so chunks never cross session boundaries. The chronological chunk stream is then grouped into RL windows of at most five chunks; a window may span sessions. The per-chunk probe pass writes the PRM items for chunk $t$. It reads the full text of chunk $t$ plus the extracted facts of chunks $t{-}1$ and $t{-}2$, used only to resolve references and decide what is new. The cross-chunk outcome pass writes ORM items from the facts of the whole window and earlier history. Structural filters and a synthetic-answer discriminability test produce the final artifact. The bottom row lists the rules that keep answer information out of the reward path (Section~\ref{sec:gold}).}
\label{fig:data}
\end{figure}

\subsection{Failure-surface taxonomy}
\label{sec:taxonomy}

Table~\ref{tab:taxonomy} lists the local failure modes. Cross-chunk questions add transition, coexistence, temporal, aggregation, multi-hop, and negative recall. This follows the capability decomposition exposed by LoCoMo and LongMemEval \citep{locomo,longmemeval}, but the questions are generated during reward construction rather than copied from those benchmarks.

\begin{table}[t]
\centering
\small
\setlength{\tabcolsep}{5pt}
\renewcommand{\arraystretch}{1.12}
\begin{tabularx}{\linewidth}{@{}p{3.05cm}Y@{}}
\toprule
Probe type & What the reward tests \\
\midrule
Specificity & Exact names, numbers, titles, versions, paths, and entities survive writing \\
Semantic precision & Negation, scope, and event meaning are not inverted \\
Plain recall & A durable explicit fact remains answerable \\
Secondary dimension & A dominant topic does not erase a smaller detail \\
Transition & The new state and the state it supersedes are both represented \\
Should-not-store & Greetings, vague praise, and generic assistant text are absent \\
Content, not meta & The memory stores the fact, not a meta-summary of it \\
Casual detail & Informal but durable personal information is retained \\
\bottomrule
\end{tabularx}
\caption{Per-chunk failure-surface probes.}
\label{tab:taxonomy}
\end{table}

\subsection{Trap probes and the precision--recall trade-off}

Recall alone makes ``store everything'' attractive. We add trap probes about content that should not have entered long-term memory, such as an assistant's generic encouragement. Their reference answer is that the bank contains no such record. If the writer stores the irrelevant statement, retrieval exposes it and the answer becomes incorrect. Traps and the bank-growth penalty are complementary. A size penalty discourages indiscriminate volume, while traps directly test semantic selectivity. A model cannot avoid the trap merely by compressing irrelevant text into a shorter entry.

\subsection{Rubrics and gold-fact attribution}
\label{sec:gold}

Each probe and outcome question contains weighted credit points with semantic equivalents; explicit full-, partial-, and zero-credit bands; named distractors, including stale or superseded values; evidence chunk ids; and, for outcome questions, the source facts introduced by each evidence chunk. The answer judge performs semantic matching; code only aggregates returned credit.

Gold facts are visible only to the outcome judge. They are not inserted into the bank, retrieval query, or answer-model prompt. Facts from earlier windows may provide context but are marked history-only and cannot generate credit for a current chunk.

\subsection{Data verification}

Reward data must be tested as an instrument, not accepted because it is well-formed JSON. Four layers are enforced. \emph{Schema validity} removes records with missing fields, unnormalized weights, illegal evidence ids, or illegal grading bands. \emph{Causal locality} requires per-chunk probes to refer to the current chunk and cross-chunk questions to contain multi-chunk evidence when required. \emph{Rubric independence} checks that credit points do not entail one another and that distractors represent real failure modes. \emph{Discriminability} checks that the judge scores five synthetic answers per item in the expected order: reference and paraphrase receive full credit, a partial answer receives partial credit, and a distractor or refusal receives zero credit. In the pilot, faithful paraphrases retain an average score of 0.90.

The structural pass keeps approximately 410k of 478k generated probes and 61k of 72k generated cross-chunk questions. To control judge cost, at most three positive probes and one trap are retained per chunk. The final artifact contains 164{,}576 positive probes, 8{,}599 traps, approximately 60.8k outcome questions, and 222{,}395 chunk-level gold-fact references.

Future work includes a random human audit with error categories and inter-annotator agreement; a train/evaluation near-duplicate and sequence-level leakage audit; per-scenario and per-failure-type calibration on strong and weak writers; and judge sensitivity to prompt, model, and repeated sampling.

\subsection{Harness-level artifact statistics}

Each complete interaction sequence is assigned to one harness configuration (Table~\ref{tab:assign}). The offline schedule preserves within-sequence window order and consumes every row exactly twice over two epochs.

\begin{table}[t]
\centering
\small
\setlength{\tabcolsep}{4pt}
\renewcommand{\arraystretch}{1.12}
\begin{tabularx}{\linewidth}{@{}p{3.45cm}YY@{}}
\toprule
Configuration & Processed units & Reward items \\
\midrule
\mbox{Hy-Memory / Chat} & 3{,}182 windows; 14{,}519 chunks & 41{,}990 recall; 2{,}769 traps; 15{,}281 ORM \\
\addlinespace[1.5pt]
\mbox{Hy-Memory / Working} & 1{,}971 windows; 8{,}902 chunks & 25{,}272 recall; 1{,}537 traps; 9{,}740 ORM \\
\addlinespace[1.5pt]
\mbox{Hy-Memory / Coding} & 1{,}154 windows; 5{,}187 chunks & 14{,}348 recall; 178 traps; 5{,}337 ORM \\
\addlinespace[1.5pt]
Mem0 & 3{,}182 windows; 14{,}516 chunks & 41{,}416 recall; 1{,}969 traps; 15{,}228 ORM \\
\addlinespace[1.5pt]
MemOS & 3{,}182 windows; 14{,}492 chunks & 41{,}550 recall; 2{,}146 traps; 15{,}241 ORM \\
\midrule
\textbf{Total} & \textbf{12{,}671 windows; 57{,}616 chunks} & \textbf{164{,}576 recall; 8{,}599 traps; $\sim$60.8k ORM} \\
\bottomrule
\end{tabularx}
\caption{Harness-level statistics of the RL reward artifact.}
\label{tab:assign}
\end{table}

\section{Reward and Credit Assignment}

\subsection{Three-hop examination}

All chunks in a window are written before scoring. For each question, the harness retrieves a memory block from the written bank, the frozen answer model answers using only that block, and the frozen judge compares the answer with the reference and weighted rubric. This scores deployable bank behavior. Directly asking a judge whether retrieved text ``contains the answer'' would not test whether contradictory or stale memories cause an incorrect response.

\subsection{Per-chunk recall reward}

For chunk $t$, let $R_t$ be positive probes and $P_t$ traps. With score $z_i\in[0,1]$,
\begin{equation}
\mathrm{prm}[t]
=\frac{\sum_{i\in R_t} z_i+\sum_{j\in P_t} z_j}{|R_t|+|P_t|}.
\end{equation}
``PRM'' here means per-chunk recall reward, not a reward model over hidden reasoning steps. A chunk with no valid probes receives zero rather than one. Items lost to an answer or judge service failure leave the denominator; policy failures remain scoreable and receive low reward.

\subsection{Attributed outcome reward}

For outcome question $q$, the judge returns $c_q[t]\in[0,1]$, the degree to which the answer recovered gold evidence introduced by chunk $t$. If $Q_t$ is the set of questions citing chunk $t$,
\begin{equation}
\mathrm{orm}[t]=\frac{1}{|Q_t|}\sum_{q\in Q_t}c_q[t].
\end{equation}
Averaging prevents a chunk from receiving more weight merely because the annotator wrote more questions about it.

\subsection{Bank-growth penalty}

Let $g_t$ be bank-token growth divided by input-chunk tokens. Each harness has a free-growth lower threshold $\ell_h$ and a full-penalty threshold $u_h$:
\begin{align}
x_t &= \mathrm{clip}\!\left(\frac{g_t-\ell_h}{u_h-\ell_h},\,0,\,1\right), \nonumber \\
\mathrm{pen}[t] &= \frac{e^{3x_t}-1}{e^3-1}.
\end{align}
Tokens are measured instead of entry count because a writer can grow a bank by lengthening existing records. Shrinking the bank is neither taxed nor directly rewarded. Figure~\ref{fig:train} labels this term Penalty.

\subsection{Combined score}

\begin{align}
s[t] &= w_{\mathrm{prm}}\sum_{k\ge t}\gamma^{k-t}\,\mathrm{prm}[k] \nonumber \\
&\quad + w_{\mathrm{orm}}\,\mathrm{orm}[t]
- w_{\mathrm{pen}}\,\mathrm{pen}[t],
\end{align}
where $w_{\mathrm{prm}}=1$, $w_{\mathrm{orm}}=2$, $w_{\mathrm{pen}}=0.3$, and $\gamma=0$. Gold-fact ORM attribution handles delayed cross-chunk evidence, so $\gamma=0$ avoids broadcasting noisy future recall scores backward.

\subsection{How this resolves sparse ORM}

Probes give every chunk a local signal, so a sequence is not limited to a few terminal questions. Weighted rubrics with partial credit and distractors replace binary answer reward. Traps and the bank-growth penalty penalize over-storage. Gold facts attribute outcome evidence to evidence chunks instead of broadcasting one score. The answer model and judge are frozen, so reader improvements cannot hide writer defects. Continuous signals increase within-group resolution. Carrying the selected policy-written bank into the next window preserves cross-session effects. This is closer in spirit to process supervision than terminal-only RL \citep{lightman}, but the supervised unit is an externally verifiable bank consequence rather than a subjective reasoning step.

\subsection{Chunk-level GRPO}

GRPO estimates a baseline from several samples without a separate value model \citep{grpo}. Standard GRPO scores each sampled trajectory once, normalizes that score within the group, and applies the resulting advantage to every token of the trajectory. A trajectory-level group is not appropriate here: all eight rollouts share the same bank only at chunk 0; after that, rollout $i$ observes the bank created by its own earlier actions, and a single trajectory score would again broadcast one judgment over all of its writes. We instead compute advantages per chunk, grouping by window, harness configuration, sequence, and chunk position:
\begin{equation}
A_{i,t}
=\frac{s_i[t]-\mu_t}{\sigma_t+\epsilon},
\end{equation}
where $\mu_t$ and $\sigma_t$ are the mean and standard deviation of $s_j[t]$ over the chunk-level group $G_t$.
The best rollout can therefore differ from chunk to chunk: rollout 1 may have the highest score at chunk 2 while rollout 3 is best at chunk 4, and each chunk's writes are reinforced or suppressed according to their own column (Figure~\ref{fig:train}).

Each rollout contributes one vote to the baseline for a chunk. If a multi-stage harness made three LLM calls for that chunk, all three segments receive the same advantage but do not count as three independent observations. Otherwise the policy could alter the baseline by changing its call count. Groups with fewer than two valid rollouts receive zero advantage.

\subsection{Bank carry-forward}

Bank selection is a separate, rollout-level decision. At the end of a window, the eight final banks are ranked by $\sum_t s[t]$. The winner becomes the initial state of the next window in that sequence; the others are discarded. If all candidates fail because of platform faults, the previous bank is reused. Bank state is checkpointed with model state.

An inherited error from an earlier optimizer step cannot receive retrospective gradient. Carrying imperfect policy-written banks is still preferable to teacher forcing with an ideal bank: deployment also requires the writer to operate on its own prior mistakes. An inherited error common to all candidates also tends to cancel in the group-relative baseline rather than being assigned to the current chunk.

\section{Training System}

Each harness is packaged behind a common interface: \texttt{bank\_open}, \texttt{write}, \texttt{search}, \texttt{bank\_stats}, and \texttt{bank\_close}. The harness owns its native control flow. Harness-invoked LLM calls are routed to the policy endpoint and recorded at generation time with token ids, rollout log-probabilities, segment boundaries, and MoE routing. This follows the harnessed-agent-RL pattern in which the deploy-time loop is kept intact while model calls connect to a trainer \citep{lightning}. Retokenizing text after execution can diverge from the sampled stream, and recomputing rollout log-probabilities with the training engine destroys the intended importance ratio.

Stage~2 starts from the OPD model, on Hy3-30B-A3B ($\sim$3B active parameters), using 16 nodes $\times$ 8 H20 GPUs (128 GPUs). A batch is 64 windows $\times$ 8 rollouts, or 512 trajectories. The objective is a PPO-style clipped surrogate with chunk-level GRPO advantages, aggregated as sequence mean then token mean, with no KL penalty. MoE routing from the rollout is replayed during training. The learning rate is $5\mathrm{e}{-6}$, constant. The context is 28{,}672 prompt tokens plus 4{,}096 response tokens. The resulting post-RL model is released and evaluated as Hy-Memory-30B-A3B. Model, optimizer, schedule, and selected banks are checkpointed every 10 steps during training.

\section{Experiments}

\subsection{Evaluation principles}

The evaluation changes only the memory writer. For each comparison, the harness version and prompts are fixed, retrieval configuration is fixed, and the answer model, judge, and other framework-side components are held fixed. Throughout this report, every reference to a DeepSeek Flash endpoint denotes \textbf{DeepSeek V4.1 Flash}. Decoding settings are matched where APIs permit, each writer builds its own bank from an empty state, and test examples and scoring rules are identical. This is necessary because memory quality is otherwise confounded with retrieval depth, reader capability, auxiliary-model capability, or a pre-existing bank.

\paragraph{Sampled benchmark matrix.}
For computational efficiency, every benchmark in the main four-harness matrix is evaluated on a fixed 20\% subsample. The evaluated post-RL model is Hy-Memory-30B-A3B. Scores are deterministic outputs of the fixed evaluation pipeline; no missing judgment is imputed as zero. Each harness score is the equal-weight average of six metrics: LoCoMo Judge, LongMemEval, PersonaMem-v2, Travel SPS, Shopping PS, and Code Memory Bench (CMB). CMB is itself the equal-weight average of two repository QA evaluations. The overall score is the equal-weight average of the four harness scores.

\subsection{Final cross-harness results}

\begin{figure}[t]
\centering
\includegraphics[width=\linewidth]{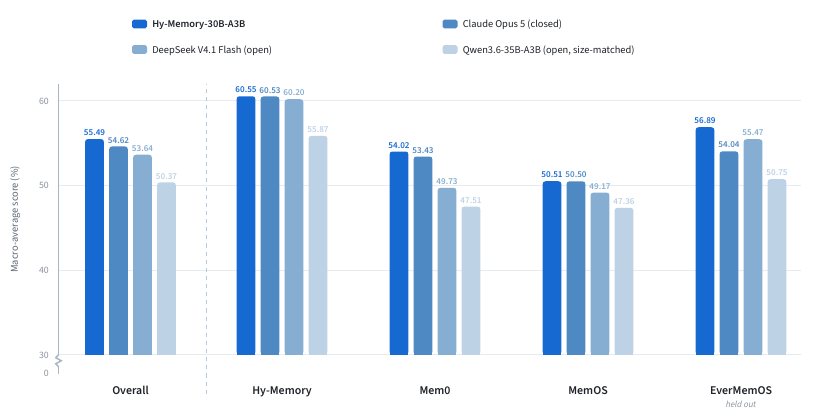}
\caption{Cross-harness macro-average scores on the fixed 20\% benchmark matrix. Each harness score averages six metrics equally; Overall averages the four harnesses equally. EverMemOS is held out from training.}
\label{fig:results}
\end{figure}

\begin{table}[t]
\centering
\small
\setlength{\tabcolsep}{3.5pt}
\begin{tabular}{@{}lrrrrr@{}}
\toprule
Memory writer & Mem0 & Hy-Mem. & MemOS & EverMem. & Avg. \\
\midrule
\textbf{Hy-Memory-30B-A3B} & \textbf{54.02} & \textbf{60.55} & \textbf{50.51} & \textbf{56.89} & \textbf{55.49} \\
Claude Opus~5 & 53.43 & 60.53 & 50.50 & 54.04 & 54.62 \\
DeepSeek V4.1 Flash & 49.73 & 60.20 & 49.17 & 55.47 & 53.64 \\
Hy3 & 45.99 & 58.06 & 50.19 & 52.30 & 51.63 \\
Qwen3.6-35B-A3B & 47.51 & 55.87 & 47.36 & 50.75 & 50.37 \\
Hy3-A3B Base & 40.20 & 50.39 & 49.05 & 50.98 & 47.66 \\
Qwen3-30B-A3B & 39.81 & 45.67 & 45.44 & 46.57 & 44.37 \\
\bottomrule
\end{tabular}
\caption{Six-metric means on fixed 20\% benchmark subsamples. The macro average weights the four harnesses equally.}
\label{tab:main-results}
\end{table}

Hy-Memory-30B-A3B ranks first on all four harnesses. Relative to Opus~5, it gains 0.60 points on Mem0, 0.02 on Hy-Memory, 0.01 on MemOS, and 2.86 on EverMemOS, for a macro-average gain of 0.87 points. Relative to DeepSeek V4.1 Flash, it gains 4.29 points on Mem0, 0.35 on Hy-Memory, 1.34 on MemOS, and 1.42 on EverMemOS, for a macro-average gain of 1.85 points. Relative to the Hy3-A3B base checkpoint, the macro average improves by 7.84 points. The 0.01--0.02 point margins on MemOS and Hy-Memory indicate parity with Opus~5 under this deterministic sampled evaluation, not a statistically resolved difference. The stronger evidence is the consistent ranking and the larger margin on the held-out framework.

The six metrics stress different behaviors, so the average improvement does not require Hy-Memory-30B-A3B to win every column. Its strongest recurring gains are on Travel SPS and CMB. On the held-out EverMemOS harness, the gain is driven particularly by Travel SPS (+10.65 points) and CMB (+13.66 points). Appendix~\ref{sec:detailed-results} reports all six benchmark scores for every model and harness.

\subsection{OOD framework generalization}

EverMemOS is not part of OPD or RL training. It is included only at evaluation time as a held-out framework and is the strongest open-source community memory baseline in this evaluation. Hy-Memory-30B-A3B reaches 56.89, compared with 55.47 for DeepSeek V4.1 Flash, 54.04 for Opus~5, and 52.30 for Hy3. This result separates model-level memory-writing skill from memorization of the training harness protocols.

\subsection{Future ablations and audits}

This report does not include controlled ablations of chunk-level versus window-level GRPO, gold-fact attribution versus broadcast ORM, positive probes versus probes plus traps, the bank-growth penalty, binary versus rubric-based rewards, RL initialization from OPD versus generic SFT, or single-harness versus mixed-harness training. We leave these comparisons to future work. Additional reward-data audits should measure human--judge agreement, sensitivity to the judge model and prompt, score distributions across synthetic answer types, separation between strong and weak memory writers, and the fraction of zero-variance rollout groups before and after reward decomposition.

\section{Engineering Lessons: Silent Failures}

Several failures looked healthy until an invariant was checked. A bank that did not persist across windows still left files on disk and produced nonzero rewards; the check is to reopen a checkpointed bank and verify a known record before training. A teacher that returned no top-$k$ values still returned successful requests, but OPD gradients were zero; missing teacher fields must raise, and teacher-token coverage must be monitored. When the per-chunk mapping was absent, a fallback broadcast one score to every segment; missing chunk-to-segment mappings must raise. A configuration omitted from a hard-coded schedule still had valid artifacts on disk; schedule manifests must be audited against the declared harness set. String rules that graded semantics produced a plausible aggregate reward; semantic conditions belong in judge rubrics, not substring code. A ``read all'' that silently paginated made the bank entry count plateau at a default limit; pagination and token accounting must be tested explicitly. Failed writes checkpointed as empty were marked complete; platform faults must be retryable and excluded from denominators. Sandboxes that shared a trajectory identity completed requests attached to the wrong rollout; every request must validate bank and trajectory identity. Malformed output treated as an environment error made bad actions disappear from training; policy mistakes should remain valid low-reward trajectories.

The governing rule is: a bad policy action must receive low reward; a platform failure must not be misrepresented as policy behavior.

\section{Limitations and Future Work}

Attribution is evidence-based rather than counterfactual: recovered gold facts do not prove that a specific write caused an answer. Credit is truncated across optimizer steps, so earlier-bank errors cannot receive retrospective gradient. One-vote grouping fixes the baseline, but multi-call chunks can still carry more loss mass. Rubric rewards reduce sparsity but require human-agreement and judge-sensitivity checks. The reward artifact covers Chat, Working, and Coding but not every long-term-memory workload; EverMemOS is only one held-out framework. The primary matrix uses deterministic 20\% subsamples, so 0.01--0.02 point margins are not statistically significant; repeated training seeds remain future work. Finally, pooled scores can hide whether gains come from extraction, update, deletion, reconciliation, or retrieval-facing representation.

\section{Conclusion}

Memory writing is a long-horizon control problem over an external state, not merely a summarization prompt. Hy-Memory-30B-A3B first learns to operate real memory harnesses through on-policy distillation, then optimizes the answerability and selectivity of the banks it creates through Memory RL. The central technical contribution is the reward interface: failure-driven probes make supervision dense, cross-chunk questions measure downstream utility, gold facts attribute that utility to source chunks, traps and growth penalties close the store-everything shortcut, and chunk-level GRPO compares actions at the correct decision boundary.

Hy-Memory-30B-A3B ranks first on Mem0, Hy-Memory, MemOS, and held-out EverMemOS. On fixed 20\% benchmark subsamples, it averages 55.49\% across six metrics and four harnesses, ahead of Claude Opus~5 at 54.62\%, DeepSeek V4.1 Flash at 53.64\%, and the Hy3-A3B base checkpoint at 47.66\%. The OOD EverMemOS result provides the clearest margin over Opus~5 (+2.86 points). Together, these results support memory writing as a transferable model capability rather than a collection of framework-specific prompt heuristics.

\section*{Resource Availability}

The Hy-Memory framework code and evaluation utilities are available at
\url{\hymemorycodeurl}. The Hy-Memory-30B-A3B model weights, configuration, and model card are
available on Hugging Face at \url{\hymemorymodelurl}.

\bibliography{references}
\bibliographystyle{colm2026_conference}

\appendix

\section{Detailed Benchmark Results}
\label{sec:detailed-results}

Tables~\ref{tab:detail-mem0}--\ref{tab:detail-evermemos} report the six benchmark scores used to compute each harness mean in Table~\ref{tab:main-results}. All scores use the same fixed 20\% benchmark subsamples as the main evaluation. Hy-Memory-30B-A3B uses the post-RL model reported throughout this paper. Checkpoint identifiers are omitted.

\newpage
\begingroup
\setlength{\abovecaptionskip}{2pt}
\setlength{\belowcaptionskip}{2pt}
\renewcommand{\arraystretch}{1.04}
\begin{minipage}{\linewidth}
\centering
\scriptsize
\setlength{\tabcolsep}{2.4pt}
\begin{tabular}{@{}lrrrrrrr@{}}
\toprule
Writer & LoCoMo & LME & PM-v2 & Travel & Shop. & CMB & Avg. \\
\midrule
\textbf{Hy-Memory-30B-A3B} & 59.54 & \textbf{73.00} & 51.62 & \textbf{54.51} & 18.33 & 67.13 & \textbf{54.02} \\
Claude Opus~5 & \textbf{69.47} & 66.00 & \textbf{53.58} & 42.98 & 17.78 & \textbf{70.74} & 53.43 \\
DeepSeek V4.1 Flash & 58.78 & 69.39 & 49.56 & 35.07 & 16.67 & 68.92 & 49.73 \\
Hy3 & 41.22 & 62.00 & 44.44 & 46.87 & 17.22 & 64.16 & 45.99 \\
Qwen3.6-35B-A3B & 50.00 & 69.00 & 49.38 & 36.39 & \textbf{20.00} & 60.30 & 47.51 \\
Hy3-A3B Base & 50.00 & 47.00 & 40.82 & 37.91 & 14.44 & 51.03 & 40.20 \\
Qwen3-30B-A3B & 46.18 & 56.00 & 42.32 & 31.98 & 16.67 & 45.68 & 39.81 \\
\bottomrule
\end{tabular}
\captionof{table}{Detailed results on the Mem0 harness.}
\label{tab:detail-mem0}
\end{minipage}
\vspace{0pt}

\begin{minipage}{\linewidth}
\centering
\scriptsize
\setlength{\tabcolsep}{2.4pt}
\begin{tabular}{@{}lrrrrrrr@{}}
\toprule
Writer & LoCoMo & LME & PM-v2 & Travel & Shop. & CMB & Avg. \\
\midrule
\textbf{Hy-Memory-30B-A3B} & 73.28 & 89.00 & 51.54 & \textbf{57.65} & 21.11 & 70.73 & \textbf{60.55} \\
Claude Opus~5 & \textbf{74.43} & \textbf{89.69} & 48.71 & 51.34 & \textbf{23.33} & 75.70 & 60.53 \\
DeepSeek V4.1 Flash & 72.90 & 85.86 & 49.00 & 53.46 & 22.78 & \textbf{77.18} & 60.20 \\
Hy3 & 69.47 & 89.00 & \textbf{52.76} & 49.01 & 21.67 & 66.44 & 58.06 \\
Qwen3.6-35B-A3B & 70.99 & 84.85 & 43.56 & 49.82 & 20.00 & 65.98 & 55.87 \\
Hy3-A3B Base & 66.41 & 87.00 & 49.84 & 31.36 & 17.78 & 49.97 & 50.39 \\
Qwen3-30B-A3B & 64.12 & 82.00 & 41.06 & 32.83 & 18.89 & 35.10 & 45.67 \\
\bottomrule
\end{tabular}
\captionof{table}{Detailed results on the Hy-Memory harness.}
\label{tab:detail-hymemory}
\end{minipage}
\vspace{0pt}

\begin{minipage}{\linewidth}
\centering
\scriptsize
\setlength{\tabcolsep}{2.4pt}
\begin{tabular}{@{}lrrrrrrr@{}}
\toprule
Writer & LoCoMo & LME & PM-v2 & Travel & Shop. & CMB & Avg. \\
\midrule
\textbf{Hy-Memory-30B-A3B} & 62.98 & 73.00 & 51.92 & 24.15 & 21.11 & 69.91 & \textbf{50.51} \\
Claude Opus~5 & \textbf{63.74} & 79.38 & \textbf{52.84} & 22.14 & 19.44 & 65.46 & 50.50 \\
DeepSeek V4.1 Flash & 61.07 & 73.00 & 50.68 & 18.61 & \textbf{21.67} & \textbf{69.97} & 49.17 \\
Hy3 & 62.98 & \textbf{81.00} & 49.64 & 18.87 & 18.89 & 69.78 & 50.19 \\
Qwen3.6-35B-A3B & 55.34 & 79.00 & 51.34 & 20.08 & 15.56 & 62.82 & 47.36 \\
Hy3-A3B Base & 58.40 & 73.00 & 50.86 & \textbf{26.05} & 21.11 & 64.88 & 49.05 \\
Qwen3-30B-A3B & 59.16 & 73.00 & 48.28 & 23.03 & 16.67 & 52.49 & 45.44 \\
\bottomrule
\end{tabular}
\captionof{table}{Detailed results on the MemOS harness.}
\label{tab:detail-memos}
\end{minipage}
\vspace{0pt}

\begin{minipage}{\linewidth}
\centering
\scriptsize
\setlength{\tabcolsep}{2.4pt}
\begin{tabular}{@{}lrrrrrrr@{}}
\toprule
Writer & LoCoMo & LME & PM-v2 & Travel & Shop. & CMB & Avg. \\
\midrule
\textbf{Hy-Memory-30B-A3B} & \textbf{73.28} & 76.00 & 53.28 & 56.56 & 19.05 & \textbf{63.19} & \textbf{56.89} \\
Claude Opus~5 & \textbf{73.28} & 80.41 & 51.19 & 45.91 & \textbf{23.89} & 49.53 & 54.04 \\
DeepSeek V4.1 Flash & 72.52 & 82.83 & 49.74 & \textbf{59.47} & 18.89 & 49.38 & 55.47 \\
Hy3 & 68.70 & \textbf{83.00} & 48.68 & 46.52 & 19.44 & 47.44 & 52.30 \\
Qwen3.6-35B-A3B & 69.08 & 79.00 & 50.60 & 43.53 & 18.77 & 43.52 & 50.75 \\
Hy3-A3B Base & 64.50 & 81.00 & 52.14 & 41.74 & 20.56 & 45.93 & 50.98 \\
Qwen3-30B-A3B & 67.94 & 60.00 & 47.10 & 44.80 & 18.33 & 41.27 & 46.57 \\
\bottomrule
\end{tabular}
\captionof{table}{Detailed results on the held-out EverMemOS harness.}
\label{tab:detail-evermemos}
\end{minipage}
\endgroup

\section{Hyperparameters}

\begingroup
\setlength{\abovecaptionskip}{3pt}
\setlength{\belowcaptionskip}{2pt}
\begin{minipage}{\linewidth}
\centering
\small
\renewcommand{\arraystretch}{1.00}
\begin{tabular}{@{}p{1.7cm}p{2.3cm}p{2.4cm}@{}}
\toprule
 & Stage 1: OPD & Stage 2: RL \\
\midrule
Model & Hy3-30B-A3B & Hy3-30B-A3B \\
Init. & Instruction-tuned & OPD step 375 \\
Harnesses & Mem0, MemOS, Hy-Memory & Same \\
Ordering & Snapshot chain & Chronological windows \\
Unit & One chunk write & One window, $\le$5 chunks \\
Rollouts & 2 per row & 8 per window \\
Signal & Top-50 reverse KL & PRM $+2$ ORM $-0.3$ Penalty \\
Advantage & Token-level & Chunk-level GRPO \\
LR & $1\mathrm{e}{-6}$ & $5\mathrm{e}{-6}$ \\
Length & 4k / 127k & 29k / 4k \\
Output & OPD model initializes RL & Hy-Memory-30B-A3B \\
\bottomrule
\end{tabular}
\captionof{table}{Training hyperparameters for the two stages.}
\end{minipage}
\endgroup

\end{document}